\documentclass[runningheads]{llncs}

\usepackage{eccv}

\usepackage{eccvabbrv}

\usepackage{graphicx}
\usepackage{booktabs}
\usepackage{algorithm}
\usepackage{algpseudocode}
\usepackage{enumitem}
\usepackage{marvosym}
\usepackage{bm}
\usepackage{multirow}
\usepackage{xcolor}
\usepackage{colortbl}
\definecolor{linecolor1}{RGB}{230, 234, 217}
\definecolor{linecolor2}{RGB}{179, 214, 159}
\definecolor{linecolor3}{RGB}{211, 222, 190}
\newcommand{\boldparagraph}[1]{\paragraph{\textbf{#1}}}

\usepackage{wrapfig}

\usepackage[accsupp]{axessibility}  

\usepackage{hyperref}

\usepackage{orcidlink}

\begin{document}

\title{LaMP: Learning Vision-Language-Action Policy
with 3D Scene Flow as \underline{La}tent \underline{M}otion \underline{P}rior}


\titlerunning{LaMP: Latent Motion As Action Prior}

\author{Xinkai Wang\inst{1,3}\orcidlink{0009-0002-7255-8372} \and
  Chenyi Wang\inst{3}\orcidlink{0009-0000-4282-9093} \and
  Yifu Xu\inst{2,3}\orcidlink{0009-0000-2122-4386} \and
  Mingzhe Ye\inst{2}\orcidlink{0009-0003-1905-3145} \and
  Fucheng Zhang\inst{3}\orcidlink{0009-0007-2610-902X}\and
  Jialin Tian\inst{2,3}\orcidlink{0009-0006-5527-3091} \and
  Xinyu Zhan\inst{2,3}\orcidlink{0009-0004-7859-2592}\and
  Lifeng Zhu\inst{1}\orcidlink{0000-0002-9999-4513}\and
  Cewu Lu\inst{2,3,4}\orcidlink{0000-0003-1533-8576}\and
  Lixin Yang\textsuperscript{\Letter}\inst{2,3}\orcidlink{0000-0001-6366-3192}
}

\authorrunning{X.~Wang et al.}

\institute{\textsuperscript{1} Southeast University, China \quad
\textsuperscript{2} Shanghai Jiao Tong University, China \\
  \textsuperscript{3} Shanghai Innovation Institute, China \quad
  \textsuperscript{4} Noematrix, China \\
\email{xinkaiwang@sii.edu.cn, siriusyang@sjtu.edu.cn}}

\renewcommand{\thefootnote}{\Letter}
\footnotetext[1]{Corresponding author.}

\maketitle

\begin{figure}[htbp]
  \centering
  \vspace{-1\baselineskip}
  \includegraphics[width=\linewidth]{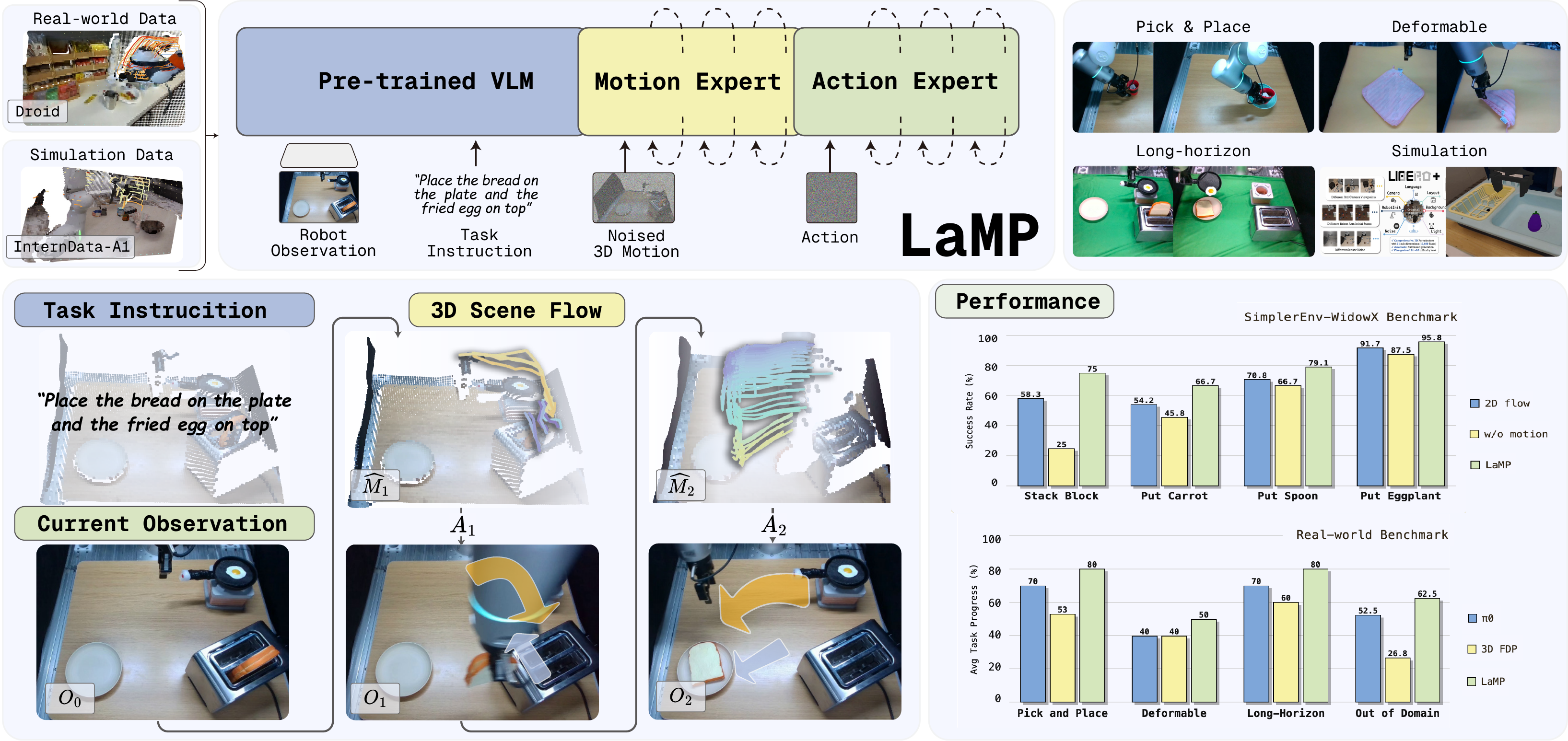}
    \caption{\textbf{LaMP: a Vision-Language-Action Model with \underline{La}tent \underline{M}otion \underline{P}rior.}
    LaMP introduces a dense 3D motion prior between VLM perception and action generation.
    The Motion Expert provides geometric foresight via one-step denoised features, which guide the Action Expert through gated cross-attention for physically grounded manipulation.
    LaMP achieves superior performance across real-world tasks and simulation benchmarks, significantly outperforming prior VLAs.}
  \label{fig:teaser}
  \vspace{-2\baselineskip}
\end{figure}

\begin{abstract}
    We introduce \textbf{LaMP}, a dual-expert Vision-Language-Action framework that embeds dense 3D scene flow as a latent motion prior for robotic manipulation.
    Existing VLA models regress actions directly from 2D semantic visual features, forcing them to learn complex 3D physical interactions implicitly.
    This implicit learning strategy degrades under unfamiliar spatial dynamics.
    LaMP addresses this limitation by aligning a flow-matching \emph{Motion Expert} with a policy-predicting \emph{Action Expert} through gated cross-attention.
    Specifically, the Motion Expert generates a one-step partially denoised 3D scene flow, and its hidden states condition the Action Expert without full multi-step reconstruction.
    We evaluate LaMP on the LIBERO, LIBERO-Plus, and SimplerEnv-WidowX simulation benchmarks as well as real-world experiments.
    LaMP consistently outperforms evaluated VLA baselines across LIBERO, LIBERO-Plus, and SimplerEnv-WidowX benchmarks, achieving the highest reported average success rates under the same training budgets. On LIBERO-Plus OOD perturbations, LaMP shows improved robustness with an average 9.7\% gain over the strongest prior baseline.
    Our project page is available at \url{https://summerwxk.github.io/lamp-project-page/}.
  \keywords{Robotic Manipulation \and 3D Scene Flow \and Vision-Language-Action Model}
\end{abstract}

\section{Introduction}
\label{sec:intro}

Humans rarely execute motor tasks by directly mapping raw perception to low-level motion commands.
Before grasping an object or closing a drawer, people implicitly form a world model of the necessary motion dynamics: which entities interact, where the interaction occurs, and how the movement unfolds.
This latent motion planning bridges the gap between high-level intentions and low-level control~\cite{thinkact,3d-vla,DDP-WM}.
Current VLA models lack an explicit intermediate representation that captures these 3D dynamic relationships, forcing them to infer contact-aware motion implicitly from action labels alone.

Existing Vision-Language-Action (VLA) models aim to equip robots with the ability to execute natural language instructions in visually rich environments~\cite{octo,openvla,gr00t,pi05,openvla-oft,eo-1,fast}.
By mapping visual and linguistic inputs to low-level motor commands, recent VLA models such as $\pi_0$~\cite{pi0}, $\pi_{0.5}$~\cite{pi05}, Gr00t~\cite{gr00t}, UniVLA~\cite{univla}, and FlowVLA~\cite{flowvla} have achieved strong generalization across tasks and embodiments.
However, a fundamental representational mismatch persists: VLM features are largely semantic and 2D-centric, whereas robot manipulation requires reasoning about explicit 3D dynamics for precise control.
This mismatch forces models to learn complex 3D physical interactions implicitly from action supervision alone, often leading to brittle execution under unfamiliar spatial dynamics.
For example, purely 2D-centric policies frequently fail on tasks requiring precise depth control (such as inserting pegs into holes or stacking objects with tight clearance) when camera viewpoints or object positions differ from training, as reported in LIBERO-Plus camera and layout perturbation benchmarks~\cite{libero-plus}.
We ask: \emph{Can dense 3D scene flow serve as a latent motion prior that explicitly grounds VLA policy learning in physical geometry?}

Prior works attempt to bridge the 2D-to-3D gap through various intermediate modalities.
FlowVLA and TraceVLA use optical flow and visual traces for temporal reasoning~\cite{flowvla,tracevla}, but 2D flow lacks explicit depth information required for contact-rich manipulation.
$\mathcal{F}_1$ and FLARE inject visual foresight into VLA control loops~\cite{f1,flare}, yet they predict pixel-space futures that conflate appearance changes with actionable kinematics.
Point-cloud diffusion policies operate directly in 3D environments~\cite{dp3,rise}, but they process geometric features as static inputs rather than modeling continuous temporal dynamics.
Even methods investigating 3D traces or flows for world modeling~\cite{tracegen,3dflowaction} generally function as standalone pipelines trained per task, without access to pretrained VLMs for semantic understanding or cross-task transfer.
Consequently, these approaches struggle to provide the continuous, scene-level spatial geometry required for robust physical interactions within a unified, language-conditioned policy.

We argue that dense 3D scene flow is not merely an observable byproduct of actions, but a fundamental latent motion prior that should actively guide policy learning.
Building on this insight, we propose \textbf{LaMP} (\textbf{La}tent \textbf{M}otion \textbf{P}rior), a dual-expert VLA framework (\cref{fig:teaser}).
LaMP aligns a flow-matching Motion Expert with a policy-predicting Action Expert through gated cross-attention.
Instead of fully reconstructing 3D scene flow (which incurs prohibitive latency), LaMP conditions action prediction on one-step partially denoised hidden motion states, inspired by predictive latent conditioning in VPP~\cite{vpp} and mimic-video~\cite{mimic_video}.
This one-step strategy avoids the computational burden of multi-step denoising while retaining task-relevant dynamics.
The gated cross-attention mechanism starts from near-zero injection and adaptively increases motion guidance only when beneficial, preventing representational collapse of pretrained VLM features while injecting 3D dynamics cues only when spatial precision is required.
This design yields actionable geometric foresight without the computational expense of full multi-step reconstruction, enabling efficient robotic deployment.

LaMP achieves the highest average success rates across LIBERO~\cite{libero}, LIBERO-Plus~\cite{libero-plus}, and SimplerEnv-WidowX~\cite{simpler-env} benchmarks, outperforming general VLAs and geometry-aware models under identical data budgets.

In summary, our contributions are as follows:
\begin{itemize}
  \item[$\bullet$] We propose \textbf{LaMP}, a dual-expert VLA framework that leverages dense 3D scene flow as a latent motion prior, bridging the gap between 2D semantic VLM features and 3D physical dynamics.
  \item[$\bullet$] We introduce a gated motion guidance mechanism that extracts partial denoised motion states and adaptively injects geometric cues into VLM representations via gated cross-attention, preventing representational collapse while maintaining computational efficiency.
  \item[$\bullet$] Extensive experiments on LIBERO, LIBERO-Plus, SimplerEnv-WidowX, and real-world tasks confirm that LaMP consistently outperforms existing VLA baselines across all benchmarks.
\end{itemize}

\section{Related Work}
\label{sec:rel_work}

\subsection{General Vision-Language-Action Models}
VLA models combine pretrained VLMs with action generation, enabling instruction-following at scale~\cite{openvla,octo,pi0,pi05,gr00t,eo-1,huang2025mobilevla}.
These approaches directly map observations to actions via end-to-end architectures, encoding scene dynamics implicitly within the policy latent space.
While achieving strong generalization on standard benchmarks, this implicit encoding often leads to compounding execution errors during long-horizon tasks.
Diffusion-based visuomotor policies~\cite{diffusion_policy} generate multimodal action sequences through iterative denoising, and point-cloud variants~\cite{dp3,rise} mitigate spatial ambiguity by operating in 3D, yet they process geometric features as static inputs rather than modeling continuous temporal dynamics.
3D-VLA~\cite{3d-vla} incorporates 3D scene understanding through point-cloud inputs, but does not follow the SimplerEnv/LIBERO evaluation protocols. We therefore treat it as related work rather than a direct end-to-end baseline.

\subsection{World Model for Visuomotor Policy}
To address the limitation of reactive policies, world model approaches predict future states before action generation.
Pixel-space methods~\cite{unipi,susie,worldvla,dreamvla} offer high expressivity but dedicate substantial capacity to reconstructing visual appearance, distracting optimization from learning actionable kinematics.
Motion-space methods provide a more control-aligned abstraction by operating directly on geometric dynamics.
$\mathcal{F}_1$~\cite{f1} and mimic-video~\cite{mimic_video} condition action synthesis on predicted future visual representations.
3DFlowAction~\cite{3dflowaction}, 3D FDP~\cite{3dFlowDiffusion}, NovaFlow~\cite{novaflow}, and TraceGen~\cite{tracegen} further ground this prediction in 3D space, using predicted flow trajectories or traces as structured motion priors~\cite{lin2026chronoflow}.
While 3D spatial priors improve physical grounding over pixel-space methods, these approaches share a common limitation: they function as standalone pipelines trained per task, without access to pretrained VLMs for semantic understanding or cross-task transfer.
LaMP takes a different position. Rather than treating 3D flow prediction as a standalone world model, we integrate it as an internal latent prior within a pretrained VLA framework, enabling embodiment-agnostic knowledge sharing and language-conditioned control with a single unified policy.
Unlike prior 3D-aware VLAs~\cite{sun2025geovla,qu2025spatialvla} that inject \emph{current-frame} 3D geometry, LaMP integrates a \emph{future, chunk-level} dense 3D scene-flow prior within a pretrained VLA framework.
Prior 3D flow methods~\cite{3dflowaction,3dFlowDiffusion,tracegen} operate as standalone pipelines without VLM semantic pretraining.
We found naive fusion of 3D features into VLM layers causes representational collapse, necessitating our gated cross-attention design.

\subsection{Intermediate Representations for Manipulation}
An alternative strategy introduces auxiliary supervision on current-frame representations rather than predicting futures~\cite{ecot}.
FlowVLA~\cite{flowvla} and TraceVLA~\cite{tracevla} use 2D optical flow and trace supervision to enhance spatial reasoning.
Qdepth-VLA, MolmoACT~\cite{qdepth,molmoact} employs depth prediction as auxiliary supervision.
UniVLA, villaX~\cite{univla,villa-x} learns latent action representations for cross-task transfer.
These approaches improve geometric awareness but operate in 2D pixel space or on static features~\cite{im2flow2act}, lacking dense temporal-spatial dynamics.
In contrast, LaMP uses dense scene-level 3D flow as a latent motion prior within the VLA framework, capturing both spatial geometry and temporal dynamics for contact-aware control.

\section{Method}
\label{sec:method}

\subsection{Problem Formulation}

We consider the problem of predicting an action chunk
$\bm{A}_{t:t+H}$ given observation $O_t$ and instruction $l$.
Let $\bm{z}_t = \phi_{\mathrm{vl}}(O_t, l)$ denote the latent
representation produced by a pre-trained vision-language
backbone. A standard VLA directly models $P(\bm{A}_{t:t+H} \mid \bm{z}_t)$.

Rather than learning a direct perception-to-action mapping,
we introduce a latent 3D motion representation $\bm{M}_t$ as an
explicit intermediate variable. The joint distribution of
actions and motion, conditioned on the current perception,
factorizes naturally by the chain rule as:
\begin{equation}
  P(\bm{A}_{t:t+H},\, \bm{M}_t \mid \bm{z}_t)
  \;=\;
  \underbrace{P(\bm{A}_{t:t+H} \mid \bm{z}_t,\, \bm{M}_t)}_{\text{Action Expert}}
  \;\;
  \underbrace{P(\bm{M}_t \mid \bm{z}_t)}_{\text{Motion Expert}}
  \label{eq:factorize}
\end{equation}
This factorization does not assume conditional independence; the Action Expert conditions on both perceptual features $\bm{z}_t$ and motion prior $\bm{M}_t$, allowing the model to leverage both semantic and dynamic information jointly.
The Motion Expert models $P(\bm{M}_t \mid \bm{z}_t)$ via
conditional flow matching, learning to generate 3D scene
flow that captures the expected scene dynamics. The
Action Expert then models
$P(\bm{A}_{t:t+H} \mid \bm{z}_t, \bm{M}_t)$, predicting robot actions
conditioned on both the perceptual features and the
motion prior.

\begin{figure*}[t]
  \centering
  \includegraphics[width=\linewidth]{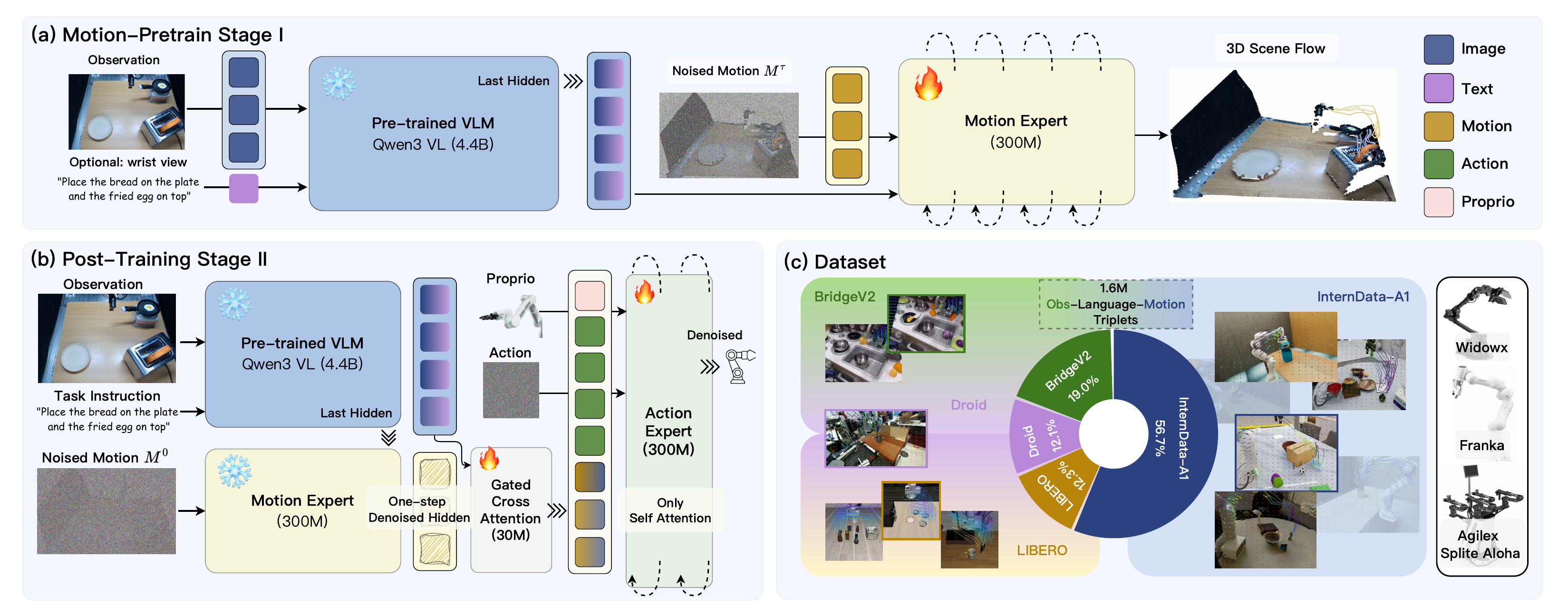}
  \caption{\textbf{Overview of LaMP.}
   (a) Motion Pre-training: The Motion Expert learns 3D scene flow prediction conditioned on VLM features. (b) Action Post-Training: Frozen Motion Expert provides one-step denoised features that fuse with VLM representations via Gated Cross-Attention. (c) Data Curation: 1.6M observation-language-motion triplets from diverse robot embodiments.}
  \label{fig:lamp_pipeline}
  \vspace{-0.5\baselineskip}
\end{figure*}

The overall pipeline is illustrated in
Figure~\ref{fig:lamp_pipeline}. 
Section~\ref{sec:method_motion} describes the Motion Expert,
Section~\ref{sec:method_action} introduces the Motion Guidance
module and the Action Expert, and
Section~\ref{sec:method_training} details the two-stage
training and inference procedures.

\subsection{Motion Expert: Learning a Latent 3D Motion Prior}
\label{sec:method_motion}

Existing intermediate representations in VLAs, such as
optical flow~\cite{flowvla,roboticvla} and visual
traces~\cite{tracevla,spatialtraces}, either remain in 2D
image space or capture only sparse point trajectories,
limiting their spatial consistency for manipulation tasks.

\boldparagraph{3D Scene Flow Representation.}
We represent the latent motion as a \emph{dense 3D scene
flow} defined over a uniform spatial grid. Specifically,
we sample a $K_h \times K_w$ grid of keypoints on the
current image plane and track each point across $T$ future
timesteps. For each keypoint at each timestep, we predict
the displacement increment
$\Delta \bm{p} = (\Delta u, \Delta v, \Delta d)$, where
$(\Delta u, \Delta v)$ denotes the 2D displacement in
image coordinates and $\Delta d$ denotes the change in depth. The full motion representation is $\bm{M}_t \in \mathbb{R}^{K \times T \times 3}$, where
$K = K_h \times K_w = 400$ and $T = 32$.
Compared to 2D flow or sparse traces, this representation
preserves scene-level 3D geometry while remaining
embodiment-agnostic, providing a strong inductive bias
for cross-embodiment generalization.
All trajectories are transformed into a unified reference camera frame $\mathrm{cam}_{\mathrm{ref}}$ to compensate for camera motion.
This screen-aligned $(u, v, d)$ representation maintains compatibility with 2D trajectory formats, enabling joint training across diverse data sources.
The dense correspondences are produced offline by the TraceForge labeler~\cite{tracegen}, which tracks each keypoint over $T{=}32$ future steps. Our 1.6M training triplets use TraceForge-provided depth, falling back to sensor depth where available (\eg, DROID~\cite{droid}). Because depth enters only at this offline labeling stage, the depth source is interchangeable and no depth sensor is required at inference.

The Motion Expert adopts a CogVideoX-style~\cite{cogvideox}
3D transformer that operates in motion-token space. We
apply spatial patchification with patch size $2 \times 2$,
grouping each $2 \times 2$ block of keypoints into a
single token, resulting in $10 \times 10$ spatial tokens
per timestep. The model is conditioned on the last-layer
VLM hidden features $\bm{z}_t$ via Adaptive LayerNorm.

We train the Motion Expert using conditional flow
matching~\cite{lipmanflow}. Given a ground-truth
motion sample $\bm{M}_t$ and Gaussian noise
$\bm{\epsilon} \sim \mathcal{N}(0, I)$, we construct the
interpolated noisy motion at time $\tau \in [0, 1]$:
\begin{equation}
  \bm{M}^\tau = (1 - \tau)\,\bm{\epsilon} + \tau\, \bm{M}_t
\end{equation}
The Motion Expert $v_\theta$ is trained to predict the
velocity field $\bm{M}_t - \bm{\epsilon}$ that transports
noise to data:
\begin{equation}
  \mathcal{L}_{\mathrm{motion}}
  = \mathbb{E}_{\tau,\, \bm{M}_t,\, \bm{\epsilon}}\!
  \left\|
  v_\theta(\bm{M}^\tau, \tau,\, \bm{z}_t)
  - (\bm{M}_t - \bm{\epsilon})
  \right\|_2^2
  \label{eq:motion_loss}
\end{equation}
\boldparagraph{One-step denoising features.}
Inspired by the partial-denoising conditioning used in video-conditioned policy learning (e.g., VPP and mimic-video~\cite{vpp,mimic_video}), the proposed LaMP framework does not require fully denoised motion latents. Given a pure noise input $\bm{M}^0 \sim \mathcal{N}(0,I)$ initialized at $\tau=0$, we simulate a single denoising step out of $N=10$ total solver steps (reaching $\tau=0.1$) and extract the intermediate motion hidden states: $\bm{z}_{\mathrm{m}}\in\mathbb{R}^{B\times N\times d_m}$.
These features retain the task-relevant dynamics while circumventing the computational cost and bias associated with forcing fully clean motion generation prior to action prediction.

\subsection{Motion Guidance and Action Expert}
\label{sec:method_action}

\boldparagraph{Single-layer motion guidance.}
Existing architectures typically perform deep fusion across multiple VLM layers. To preserve the pretrained VLM representations while still injecting temporal dynamics, we adapt gated cross-attention to fuse one-step partially denoised motion features into the last-layer VLM features. Let $\tilde{\bm{h}}_{\mathrm{motion}}=\mathrm{LN}(\bm{W}_{\mathrm{proj}}\,\bm{z}_{\mathrm{m}})$, where $\mathrm{LN}(\cdot)$ denotes layer normalization:
\begin{equation}
  \bm{z}_{\mathrm{guided}}
  =
  \bm{z}
  +\sigma(g)\cdot
  \mathrm{CA}\!\left(
    Q=\mathrm{LN}(\bm{z}),\,
    K=V=\tilde{\bm{h}}_{\mathrm{motion}}
  \right)
\end{equation}
where $\mathrm{CA}$ represents the multi-head cross-attention operation, and $g$ is a learnable scalar gate initialized at zero, shared across all attention heads and spatial positions.
We apply the sigmoid activation $\sigma(\cdot)$ to constrain the gate value to $(0, 1)$, allowing the model to start from a weak motion-injection regime and increase the guidance only when it proves beneficial, leading to more stable optimization.

\boldparagraph{Action expert.}
The Action Expert receives the guided features $\bm{z}_{\mathrm{guided}}$ and predicts continuous actions via flow matching. Let $\bm{a}^\tau=(1-\tau)\bm{\epsilon}^a+\tau \bm{a}$, where $\bm{\epsilon}^a\sim\mathcal{N}(0,I)$. The action objective is given by
\begin{equation}
  \mathcal{L}_{\mathrm{action}}
  =
  \mathbb{E}_{\tau,\bm{a},\bm{\epsilon}^a}
  \left\|
  v_\phi(\bm{a}^\tau,\tau,\bm{z}_{\mathrm{guided}})
  -(\bm{a}-\bm{\epsilon}^a)
  \right\|_2^2
\end{equation}
Utilizing motion-guided features as the condition ensures that action denoising depends on explicit 3D dynamics rather than solely on 2D implicit semantics.

\subsection{Two-Stage Training and Inference}
\label{sec:method_training}

\boldparagraph{Stage 1: motion-prior pretraining.}
We train the Motion Expert using the objective $\mathcal{L}_{\mathrm{motion}}$ on dense 3D scene-flow supervision generated via the TraceForge pipeline~\cite{tracegen}. Using this method, we construct 1.6M observation-language-motion triplets from the LIBERO~\cite{libero}, BridgeV2~\cite{bridgev2}, DROID~\cite{droid}, and InternData-A1~\cite{interndata-a1} datasets.

\boldparagraph{Stage 2: motion-guided policy learning.}
We freeze the pretrained Motion Expert and the VLM backbone, and subsequently train the Motion Guidance module and the Action Expert. The optimization in this stage employs solely the objective $\mathcal{L}_{\mathrm{action}}$. Crucially, we obtain the guidance features by performing the exact same partial denoising process used during inference: starting from pure noise $\bm{m}^0 \sim \mathcal{N}(0,I)$, we run the Motion Expert for one step of a 10-step ODE solver and extract the resulting hidden state.

\boldparagraph{Inference.}
During inference, LaMP performs four sequential steps:
(1) encode the RGB observation $o_t$ and the instruction $l$ using the VLM and read the last-layer features;
(2) execute the Motion Expert starting from Gaussian noise at $\tau=0$ for one step (out of $N=10$) to $\tau=0.1$ and extract the motion hidden state $\bm{z}_{\mathrm{m}}$;
(3) fuse the motion information into the VLM features via a single-layer gated cross-attention;
(4) denoise in the action space from $\tau=0$ to $\tau=1$ to sample the clean action $\bm{a}_t$.
The resulting policy operates online and does not require full motion reconstruction during inference. 
\section{Simulation Experiments}
\label{sec:sim_exp}

We design our empirical evaluation to answer the following research questions:
\begin{itemize}[noitemsep,leftmargin=*]
  \item[$\bullet$] \textbf{Q1: Geometric Foresight.} Does explicitly modeling future 3D scene flow as a generative action prior improve performance on long-horizon planning and fine-grained spatial reasoning tasks? (see~\cref{table:main_comparison} and~\cref{sec:ablation})
  \item[$\bullet$] \textbf{Q2: 3D Geometry vs.\ 2D Pixels.} Is depth-aware 3D scene flow superior to 2D optical flow for guiding manipulation policies? (see~\cref{fig:simpler_ablation})
  \item[$\bullet$] \textbf{Q3: Out-of-Distribution Robustness.} Does the 3D motion prior facilitate robust zero-shot generalization under visual perturbations and cross-domain transfer? (see~\cref{table:main_comparison} and~\cref{table:libero_plus_comparison})
  \item[$\bullet$] \textbf{Q4: Fusion Strategy.} Does gated cross-attention outperform simpler feature fusion strategies for integrating the motion prior into the action policy? (see~\cref{fig:gated_ablation})
\end{itemize}

\subsection{Evaluation Setup}

We evaluate our method on two standard simulation environments and three public benchmarks. \textbf{LIBERO} and \textbf{LIBERO-Plus}~\cite{libero, libero-plus} assesses language-conditioned manipulation, spatial reasoning, and long-horizon planning under object and layout variations. \textbf{SimplerEnv-WidowX}~\cite{simpler-env} serves as a high-fidelity proxy for real-world robustness, evaluating policies under realistic visual mismatches.

\boldparagraph{Training Recipe.}
Following our decoupled two-stage paradigm, the Motion Expert is pretrained on TraceForge-generated~\cite{tracegen} dense 3D scene-flow supervision derived from a mixture of datasets. Subsequently, the Motion Expert and the Qwen3-VL-4B-Instruct backbone are frozen, and we train the Action Expert together with the motion-guidance module. Unless otherwise specified, LaMP employs the one-step partial denoising strategy for motion guidance during both training and inference. 
Our training pipeline is built upon the starVLA framework~\cite{starvla2025} and all experiments are conducted on 16 NVIDIA H100 GPUs. Detailed hyperparameters, such as the learning rate and batch size, are provided in the supplementary material.
At inference, the Motion Expert adds only $1.35\times$ latency and $1.15\times$ memory while sustaining real-time control (App.~\ref{appendix:runtime}).

\boldparagraph{LIBERO Benchmark.} We evaluate LaMP on all four standard suites: LIBERO-Spatial, LIBERO-Object, LIBERO-Goal, and LIBERO-Long, comprising 10 tasks per suite with 500 expert demonstrations each, while training a single policy across all suites.
We compare LaMP against a diverse set of VLA architectures: General VLAs (OpenVLA~\cite{openvla}, OpenVLA-OFT~\cite{openvla-oft}, $\pi_0$~\cite{pi0}, $\pi_{0.5}$~\cite{pi05}, GR00T N1~\cite{gr00t}), latent-action VLAs (UniVLA~\cite{univla}, villa-X~\cite{villa-x}),  video-based VLAs (mimic-video~\cite{mimic_video}, WorldVLA~\cite{worldvla}, \ensuremath{\mathcal{F}_1}~\cite{f1}), and 2D flow/trace-guided VLAs (FlowVLA~\cite{flowvla}, TraceVLA~\cite{tracevla}) and report the task success rate for each suite and the average across all tasks. Each task is evaluated over 50 episodes.

\begin{table*}[!tbp]
\vspace{-1\baselineskip}
  \caption{\textbf{Comparison of different methods on the LIBERO and SimplerEnv-WidowX benchmarks.} We report the task success rate for each suite and the average. \textbf{Bold} denotes the best performance, and \underline{underline} denotes the second best.}
  \label{table:main_comparison}
  \vspace{-1\baselineskip}
  \begin{center}
    \begin{small}
      \setlength{\tabcolsep}{3pt}
      \resizebox{\textwidth}{!}{
      \begin{tabular}{l|ccccc|ccccc}
        \toprule
        \multirow{2}{*}{Method} & \multicolumn{5}{c|}{\textbf{LIBERO}} & \multicolumn{5}{c}{\textbf{SimplerEnv-WidowX}} \\
        \cmidrule(lr){2-6} \cmidrule(lr){7-11}
        & Spatial & Object & Goal & Long & \textbf{Avg} & Stack Block & Put Carrot & Put Spoon & Put Eggplant & \textbf{Avg} \\
        \midrule
        \multicolumn{11}{l}{\textit{\textbf{General VLA}}} \\
        OpenVLA~\cite{openvla} & 84.7 & 88.4 & 79.2 & 53.7 & 76.5 & 0.0 & 0.0 & 4.2 & 12.5 & 4.2 \\
        OpenVLA-OFT~\cite{openvla-oft} & 97.6 & 98.4 & \cellcolor{linecolor1}{\underline{97.9}} & 94.5 & 97.1 & -- & -- & -- & -- & -- \\
        $\pi_0$~\cite{pi0} & 96.8 & 98.8 & 95.8 & 85.2 & 94.2 & 16.7 & 0.0 & 29.1 & 62.5 & 40.1 \\
        $\pi_{0.5}$~\cite{pi05} & \cellcolor{linecolor1}{\underline{98.8}} & 98.2 & \cellcolor{linecolor2}{\textbf{98.0}} & 92.4 & 96.9 & 44.7 & 64.7 & 49.3 & 69.7 & 57.1 \\
        GR00T N1~\cite{gr00t} & 94.4 & 97.6 & 93.0 & 90.6 & 93.9 & 16.7 & 45.8 & 62.5 & 20.8 & 49.5 \\
        \midrule
        \multicolumn{11}{l}{\textit{\textbf{Latent-Action VLA}}} \\
        UniVLA~\cite{univla} & 96.5 & 96.8 & 95.6 & 92.0 & 95.2 & 29.2 & 62.5 & \cellcolor{linecolor2}{\textbf{83.3}} & \cellcolor{linecolor2}{\textbf{100.0}} & 68.7 \\
        villa-X~\cite{villa-x} & 97.5 & 97.0 & 91.5 & 74.5 & 90.1 & 61.3 & 46.3 & 77.9 & 64.6 & 62.5 \\
        \midrule
        \multicolumn{11}{l}{\textit{\textbf{Video-Based VLA}}} \\
        mimic-video~\cite{mimic_video} & 94.2 & 96.8 & 90.6 & -- & 93.9 & 29.2 & 54.2 & 41.7 & \cellcolor{linecolor2}{\textbf{100.0}} & 56.3 \\
        WorldVLA~\cite{worldvla} & 87.6 & 96.2 & 83.4 & 60.0 & 81.8 & -- & -- & -- & -- & -- \\
        \ensuremath{\mathcal{F}_1}~\cite{f1} & 98.2 & 97.8 & 95.4 & 91.3 & 95.7 & 50.0 & \cellcolor{linecolor2}{\textbf{70.8}} & 50.0 & 66.7 & 72.9 \\  
        \midrule
        \multicolumn{11}{l}{\textit{\textbf{2D Flow/Trace-Guided VLA}}} \\
        FlowVLA~\cite{flowvla} & 93.2 & 95.0 & 91.6 & 72.6 & 88.1 & \cellcolor{linecolor1}{\underline{62.5}} & 62.5 & 70.8 & \cellcolor{linecolor2}{\textbf{100.0}} & \cellcolor{linecolor1}{\underline{74.0}} \\
        TraceVLA~\cite{tracevla} & 84.6 & 85.2 & 75.1 & 54.1 & 75.8 & 16.6 & 16.6 & 12.5 & 65.0 & 27.7 \\
        \midrule
        \multicolumn{11}{l}{\textit{\textbf{3D-Aware VLA}}} \\
        GeoVLA~\cite{sun2025geovla} & 98.4 & \cellcolor{linecolor1}{\underline{99.0}} & 96.6 & \cellcolor{linecolor1}{\underline{96.6}} & \cellcolor{linecolor1}{\underline{97.7}} & -- & -- & -- & -- & -- \\
        SpatialVLA~\cite{qu2025spatialvla} & 88.2 & 89.9 & 78.6 & 55.5 & 78.1 & 29.2 & 25.0 & 16.7 & \cellcolor{linecolor2}{\textbf{100.0}} & 42.7 \\
        \midrule
        \textbf{LaMP} & \cellcolor{linecolor2}{\textbf{99.4}} & \cellcolor{linecolor2}{\textbf{99.8}} & 97.4 & \cellcolor{linecolor2}{\textbf{96.7}} & \cellcolor{linecolor2}{\textbf{98.3}} & \cellcolor{linecolor2}{\textbf{75.0}} & \cellcolor{linecolor1}{\underline{66.7}} & \cellcolor{linecolor1}{\underline{79.1}} & \cellcolor{linecolor1}{\underline{95.8}} & \cellcolor{linecolor2}{\textbf{79.2}} \\
        \hspace*{0.5cm} w/o \emph{motion} & 95.8 & 98.9 & 96.6 & 78.2 & 92.4 & 25.0 & 45.8 & 66.7 & 87.5 & 56.3 \\
        \bottomrule
      \end{tabular}
      }
    \end{small}
  \end{center}
  \vspace{-1\baselineskip}
\end{table*}

As reported in Table~\ref{table:main_comparison}, LaMP consistently outperforms strong baselines on LIBERO. In particular, LaMP achieves the best average success rate (98.3\%) and the best long-horizon performance on LIBERO-Long (96.7\%), exceeding the strongest prior baseline $\pi_{0.5}$ (96.9\% average, 92.4\% on Long). These results confirm that explicit 3D motion modeling reduces error accumulation over long manipulation horizons, directly addressing \textbf{Q1}.

\boldparagraph{LIBERO-Plus Benchmark (Zero-Shot OOD).}
All methods in Table~\ref{table:libero_plus_comparison} are trained exclusively on the original LIBERO demonstrations and evaluated zero-shot on the seven perturbation dimensions of LIBERO-Plus~\cite{libero-plus}, without any additional training data.
LaMP achieves the highest average success rate (79.3\%), outperforming the strongest prior baseline OpenVLA-OFT (69.6\%) by 9.7 percentage points.
The gains concentrate on perturbations that alter the visual appearance or spatial layout of the scene: LaMP reaches 97.4\% on Background and 95.3\% on Light, surpassing all baselines by clear margins. On Robot perturbation, which changes the arm's kinematic configuration, LaMP attains 69.6\% while the next best method ($\pi_0$-Fast) reaches only 21.6\%. This robustness stems from our camera-frame motion prior: by reasoning in $(\Delta u,\Delta v,\Delta d)$ space rather than joint space, the learned dynamics transfer across visual and kinematic variations without task-specific adaptation.
Ablating the Motion Expert (w/o \emph{motion expert}) reduces the average from 79.3\% to 71.6\%, with the largest drops on Camera ($-$17.8) and Robot ($-$13.6), confirming that the motion prior is the primary source of OOD robustness rather than the VLM backbone alone, directly addressing \textbf{Q3}.

\begin{table*}[t]
  \caption{\textbf{LIBERO-Plus zero-shot OOD evaluation.} 
All models are trained on LIBERO and evaluated zero-shot on seven perturbation dimensions without additional training data.}
  \vspace{-1\baselineskip}
  \label{table:libero_plus_comparison}
  \begin{center}
    \begin{small}
    \setlength{\tabcolsep}{3pt}
    \resizebox{0.9\columnwidth}{!}{
      \begin{tabular}{lcccccccc}
        \toprule
        \multirow{2}{*}{Method} & \multicolumn{8}{c}{\textbf{LIBERO-Plus}}\\
        \cmidrule(lr){2-9}
        & Camera & Robot & Language & Light &Background &Noise &Layout &Avg\\
        \midrule
        UniVLA~\cite{univla} & 1.8 & 46.2 & 69.6 & 69.0 &81.0 &21.2 &31.9 &42.9 \\
        OpenVLA~\cite{openvla} & 0.8 & 3.5 & 23 & 8.1 & 34.8 & 15.2 & 28.5 & 15.6 \\
        OpenVLA-OFT~\cite{openvla-oft} & 56.4 & 31.9 & 79.5 & \cellcolor{linecolor1}{\underline{88.7}} & 93.3 & 75.8 & \cellcolor{linecolor2}{\underline{\textbf{74.2}}} & 69.6 \\
        $\pi_0$~\cite{pi0} & 13.8 &6.0 &58.8 &85.0 &81.4 &\cellcolor{linecolor2}{\textbf{79.0}}& 68.9 &53.6 \\
        $\pi_0 \texttt{-Fast} $~\cite{fast} & \cellcolor{linecolor2}{\textbf{65.1}} &21.6 &61.0 &73.2 &73.2 &74.4 &68.8 &61.6\\
        WorldVLA~\cite{worldvla} & 0.1& 27.9& 41.6& 43.7 &17.1& 10.9& 38.0& 25.0 \\
        \midrule
        \textbf{LaMP} & \cellcolor{linecolor1}{\underline{64.5}} & \cellcolor{linecolor2}{\textbf{69.6}} & \cellcolor{linecolor2}{\textbf{88.2}} & \cellcolor{linecolor2}{\textbf{95.3}} & \cellcolor{linecolor2}{\textbf{97.4}} & \cellcolor{linecolor1}{\underline{76.9}} & \cellcolor{linecolor1}{\underline{73.8}} & \cellcolor{linecolor2}{\textbf{79.3}} \\
        \hspace*{0.5cm} w/o \emph{motion} & 46.7 & \cellcolor{linecolor1}{\underline{56.0}} & \cellcolor{linecolor1}{\underline{82.5}} & \cellcolor{linecolor2}{\textbf{95.3}} & \cellcolor{linecolor1}{\underline{95.4}} & 69.3 & 71.0 & \cellcolor{linecolor1}{\underline{71.6}}\\
        \bottomrule
      \end{tabular}
      }
      \vspace{-1\baselineskip}
    \end{small}
  \end{center}
\end{table*}

\boldparagraph{SimplerEnv-WidowX Benchmark.}
Following the SimplerEnv protocol~\cite{simpler-env}, we evaluate policies 
trained on BridgeV2 real-world data without any fine-tuning on the 
simulated WidowX embodiment. LaMP achieves \textbf{79.2\%} average success 
rate, outperforming the strongest baseline FlowVLA (74.0\%) by \textbf{5.2\%} 
and $\pi_{0.5}$ (57.1\%) by \textbf{22.1\%} 
(Table~\ref{table:main_comparison}). Notably, on Stack Block, the most 
challenging task where all prior methods struggle, LaMP reaches 
\textbf{75.0\%} while the second-best FlowVLA achieves only 62.5\% and
$\pi_{0.5}$ drops to 44.7\%.
LaMP also clearly surpasses recent 3D/depth-aware VLAs in this cross-domain setting, \eg, SpatialVLA (42.7\%) and RoboVLMs (38.0\%)~\cite{qu2025spatialvla,li2026robovlms}.

This strong sim-to-sim transfer despite the visual domain gap stems from 
our 3D scene-flow prior. By reasoning in 3D camera coordinate space rather than pixel space, LaMP acquires a geometrically 
grounded representation that is robust to visual appearance changes
between real and simulated environments, further addressing \textbf{Q3}.
LaMP also generalizes to a bimanual embodiment, reaching $83.9/81.8$ average success on 10 RoboTwin 2.0 tasks~\cite{chen2025robotwin} under clean/randomized settings (App.~\ref{appendix:robotwin}).

\subsection{Ablation Studies}
\label{sec:ablation}

\begin{wrapfigure}[16]{r}{0.5\textwidth}
\vspace{-2.0\baselineskip}
  \begin{center}
    \includegraphics[width=0.5\textwidth]{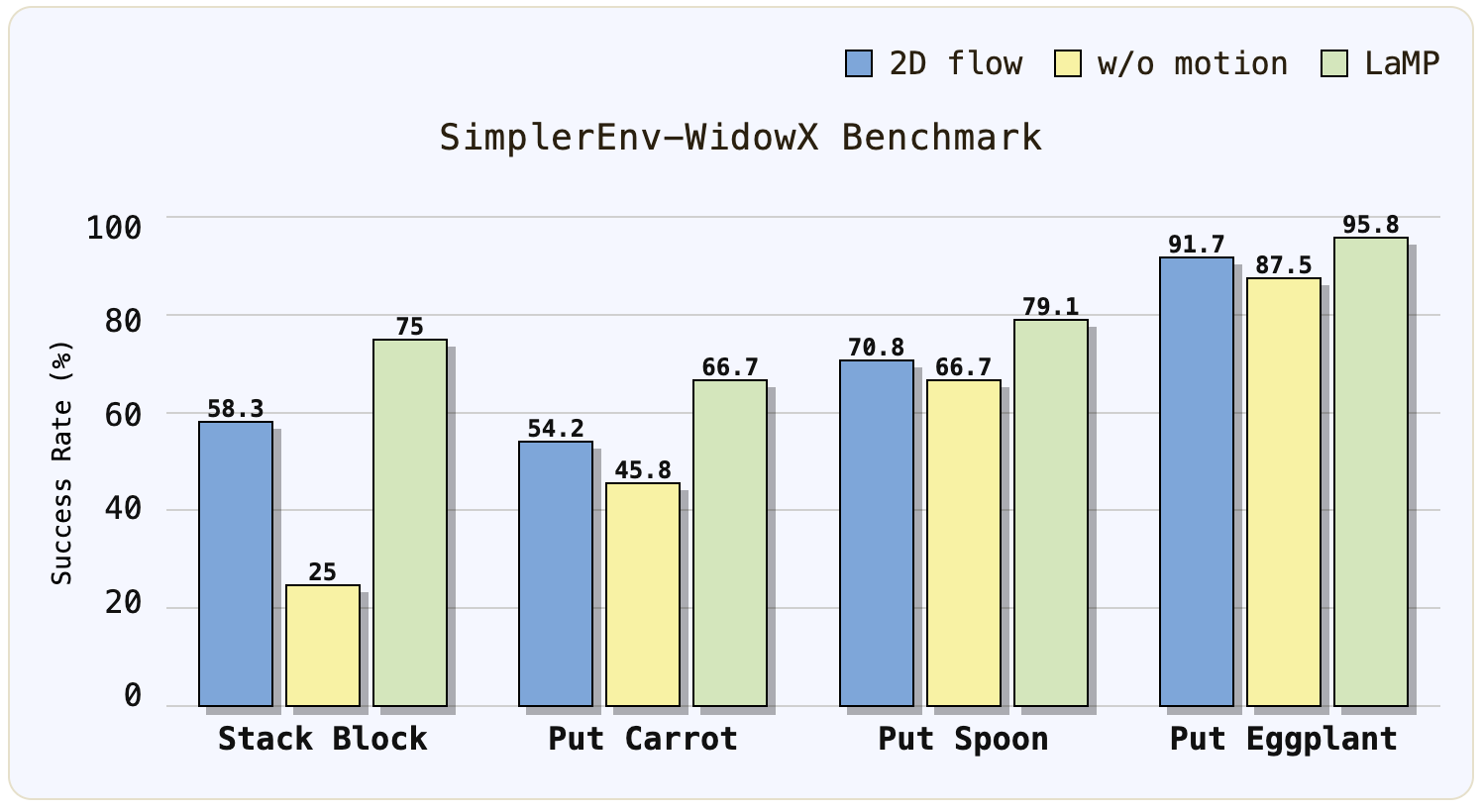}
  \end{center}
  \vspace{-1\baselineskip}
  \caption{
  \textbf{Ablation study on the SimplerEnv-WidowX benchmark.} Success rates of LaMP are compared against variants without motion priors and with 2D flow priors across four manipulation tasks.}
  \label{fig:simpler_ablation}
\vspace{-1\baselineskip}
\end{wrapfigure}

\begin{wrapfigure}[15]{r}{0.5\textwidth}
  \begin{center}
    \includegraphics[width=0.5\textwidth]{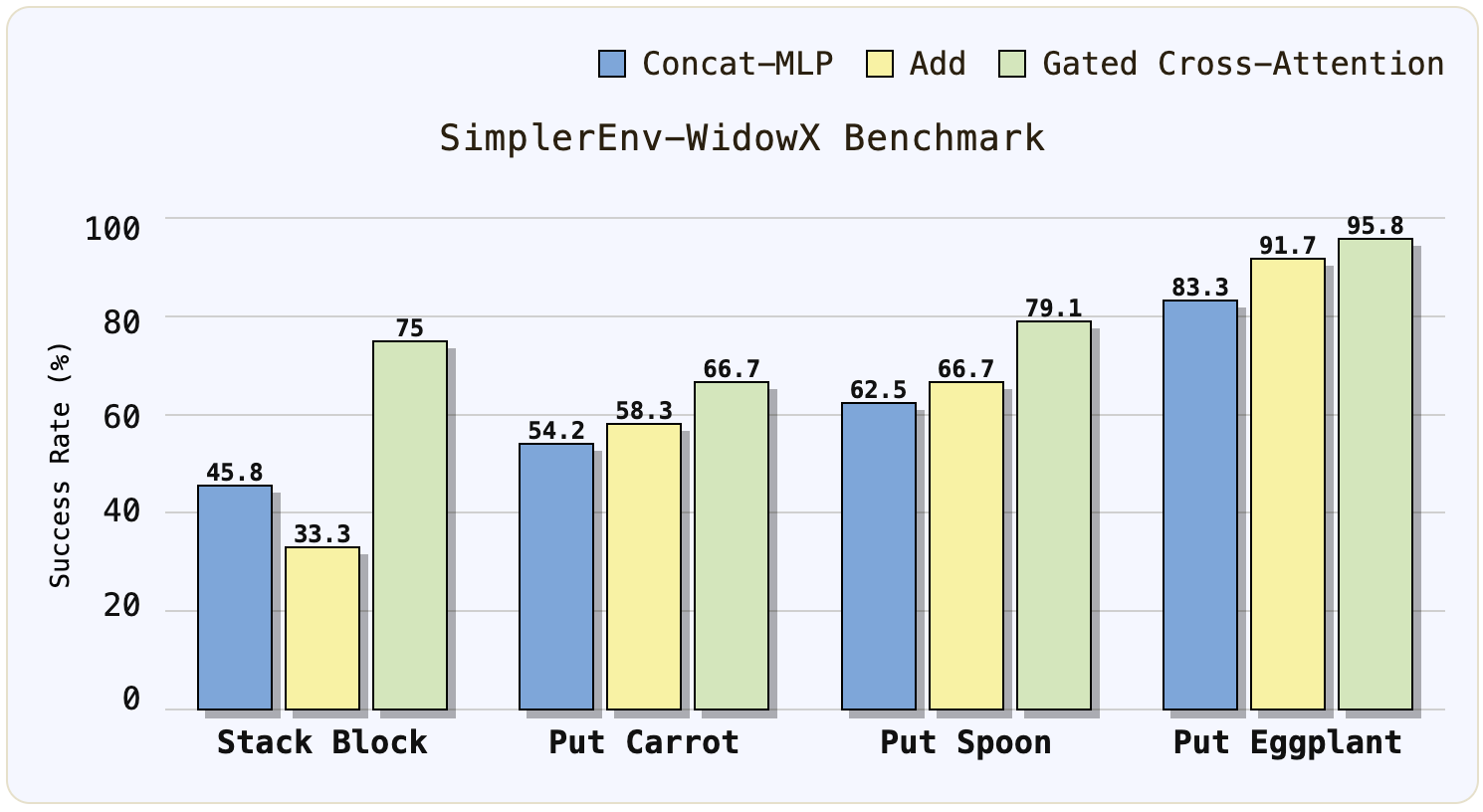}
  \end{center}
  \vspace{-1\baselineskip}
  \caption{
  \textbf{Ablation study on fusion strategies.} Success rates of Gated Cross-Attention are compared against Concat-MLP and Add variants across four manipulation tasks.}
  \label{fig:gated_ablation}
\end{wrapfigure}

To systematically address our RQs and isolate the contributions of our motion prior and architectural choices, we evaluate four model variants on SimplerEnv-WidowX benchmark (see~\cref{fig:simpler_ablation} and~\cref{fig:gated_ablation}):
\begin{enumerate}
  \item \textbf{No-Motion}. The Motion Expert is entirely removed, reducing the model to a purely reactive VLA without generative foresight.
  \item \textbf{2D-Flow Prior}. The dense 3D scene flow prior is replaced with a 2D optical-flow prior by masking out the depth dimension during training while keeping $u$ and $v$ unchanged.
  \item \textbf{Add}. The denoised motion hidden states are directly added to the corresponding action tokens at each layer, without any learnable gating.
  \item \textbf{Concat+MLP}. The motion hidden states and action tokens are concatenated along the feature dimension and projected back via a two-layer MLP.
\end{enumerate}

\boldparagraph{Effectiveness of Geometric Foresight.}
Comparing LaMP with the No-Motion variant reveals severe performance degradation across both benchmarks. On LIBERO (See~\cref{table:main_comparison}), removing the Motion Expert reduces the average success rate from 98.3\% to 92.4\% ($-$5.9\%), with the largest drop on LIBERO-Long (96.7\% $\rightarrow$ 78.2\%, $-$18.5\%). On SimplerEnv-WidowX (See~\cref{fig:simpler_ablation}), the degradation is even more pronounced: the average drops from 79.2\% to 56.3\% ($-$22.9\%), with Stack Block plummeting by 50.0\% (75.0\% $\rightarrow$ 25.0\%). This substantial gap confirms that explicit visual foresight is critical for long-horizon planning and bridging the real-to-sim domain gap. Without generating future geometric states, the policy struggles with goal alignment in complex multi-stage scenes and fails to transfer skills learned from BridgeV2 demonstrations to unseen visual configurations in simulation, directly addressing \textbf{Q1}.

\boldparagraph{3D Geometry vs.\ 2D Pixels.}
Compared to the 2D-Flow variant, our 3D prior yields consistent gains across all four tasks. While 2D flow provides temporal cues, it lacks physical grounding for precise spatial manipulation. The 3D scene-flow prior improves Stack Block by 16.7\% (58.3\% $\rightarrow$ 75.0\%) and Put Spoon by 8.3\% (70.8\% $\rightarrow$ 79.1\%). Notably, on Put Eggplant both methods achieve comparable performance (91.7\% vs.\ 95.8\%), suggesting that tasks requiring less fine-grained 3D reasoning can be solved with 2D temporal cues alone. This result confirms that depth-aware geometric representations are indispensable for contact-rich manipulation tasks with stringent spatial constraints, directly addressing \textbf{Q2}.

\boldparagraph{Gated Fusion vs.\ Alternative Fusion Strategies.}
To investigate how the motion prior is best integrated into the action policy, we compare our gated cross-attention module against the Add and Concat+MLP variants described above on the SimplerEnv-WidowX benchmark. 

As shown in~\cref{fig:gated_ablation}, gated cross-attention consistently outperforms both alternatives across all four tasks. The most striking gap appears on Stack Block, where the \textbf{Add} variant achieves only 33.3\% compared to 75.0\% for our gated module ($-$41.7\%), and \textbf{Concat+MLP} reaches 45.8\% ($-$29.2\%). This collapse indicates that unweighted motion injection harms performance when the predicted flow conflicts with visual observations. On simpler tasks such as Put Eggplant, the performance gap narrows (\textbf{Add}: 91.7\%, \textbf{Concat+MLP}: 83.3\%, \textbf{Gated}: 95.8\%), confirming that selective gating is most critical when spatial precision is required. These results show that the gated mechanism is essential for reliably exploiting the motion prior across tasks of varying difficulty, directly addressing \textbf{Q4}.

\section{Real-World Experiments}
\label{sec:real_exp}
\subsection{Setup}
\begin{figure*}[t]
  \centering
  \includegraphics[width=\linewidth]{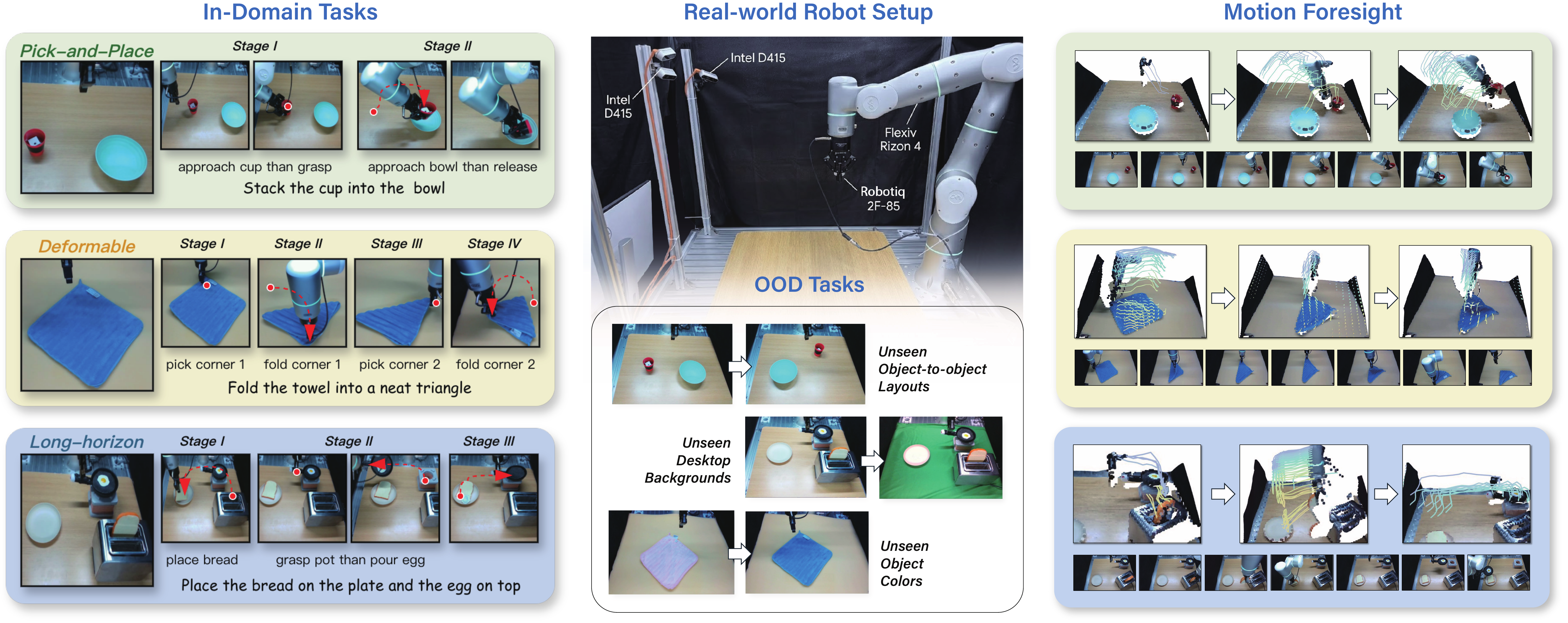}
  \caption{\textbf{Real-world experiment platform and task overview.}
\textbf{(a) In-domain tasks:} Pick-and-Place, Deformable manipulation, and Long-horizon tasks.
\textbf{(b) top: Hardware setup:} Flexiv Rizon 4 arm with Robotiq 2F-85 gripper and Intel D415 camera. \textbf{bottom: OOD test conditions:} unseen layout, object, and background.
\textbf{(c) Visualizations of motion foresight.} The predicted motion trajectories are overlaid on the current observation. The color gradient from blue to red indicates the temporal progression of the predicted motion.}
  \label{fig:platform}
  \vspace{-1\baselineskip}
\end{figure*}

\boldparagraph{Hardware Platform.}
We conduct real-world experiments using a 7-DoF Flexiv Rizon 4 robotic arm equipped with a 1-DoF Robotiq 2F-85 gripper. A single Intel RealSense D415 RGB-D camera is mounted to provide global visual observations.
The action space consists of a 6-DoF end-effector pose (relative to the current pose) and a 1-DoF gripper open/close state.
All devices are connected to a workstation equipped with an NVIDIA RTX 4090 GPU for model inference and real-time control.
Robot demonstrations are collected via haptic teleoperation.

\subsection{Evaluation}
\boldparagraph{In-Domain Tasks.}
We design three representative manipulation tasks spanning rigid-body manipulation, deformable-object folding, and long-horizon multi-step planning, as illustrated in \cref{fig:platform} (left).

\textit{Pick-and-Place (Stack Cup)} is a rigid-body 6-DoF pick-and-place task in which the robot grasps a blue cup and stacks it into a green bowl.
The task proceeds in two stages:
(I)~approach the cup from the side and grasp it, and (II)~approach the bowl and release the cup inside.
Successful execution requires precise 6-DoF positioning to avoid collisions with the bowl rim, thereby evaluating the spatial reasoning capability of the policy for contact-aware trajectory planning.

\textit{Fold Towel} requires the robot to fold a blue towel into a neat triangle through sequential corner manipulation. The task consists of four stages:
(I)~pick corner~1, (II)~fold corner~1 to the opposite edge,
(III)~pick corner~2, and (IV)~fold corner~2 to complete the triangle.
The task is considered successful if the resulting shape closely approximates a triangle with well-aligned edges.
This task evaluates the ability of the policy to handle deformable dynamics and sequential multi-step reasoning.

\textit{Making Bread} is a long-horizon multi-step task that requires the robot to serve bread and pour an egg.
The task involves three stages: 
(I)~place the bread onto a plate, (II)~grasp the handle of the pot and pour the egg onto the bread, and (III)~return the pot to its original position.
This task evaluates the capability of the policy for long-horizon planning and maintaining task coherence across multiple manipulation primitives.

\boldparagraph{Out-of-Distribution (OOD) Tasks.}
To evaluate generalization beyond the training distribution, we
construct three conditions unseen in the expert demonstrations,
as shown in \cref{fig:platform} (middle).

\textit{Unseen Layout}:  positions and orientations in the workspace are significantly altered from those observed during training, testing the robustness of the policy to spatial configuration changes.

\textit{Unseen Object}: Object instances are replaced with unseen
counterparts that are not present in the training demonstrations. Specifically,
for \textit{Pick-and-Place}, the blue cup is replaced with a red
cup, or the green bowl is replaced with a white bowl, testing
generalization to unseen colors. For \textit{Fold Towel},
evaluation is conducted using an unseen towel of a different color. For \textit{Making Bread}, the plate used for serving
is replaced with a bowl, which requires the policy to adapt to a
different container geometry.

\textit{Unseen Background}: The workspace surface is changed
(\eg, replacing the wooden tabletop with a green mat), introducing
background variations not covered by the expert demonstrations
and evaluating the visual robustness of the policy.

\begin{wrapfigure}{r}{0.5\textwidth}
\vspace{-2\baselineskip}
  \begin{center}
    \includegraphics[width=0.5\textwidth]{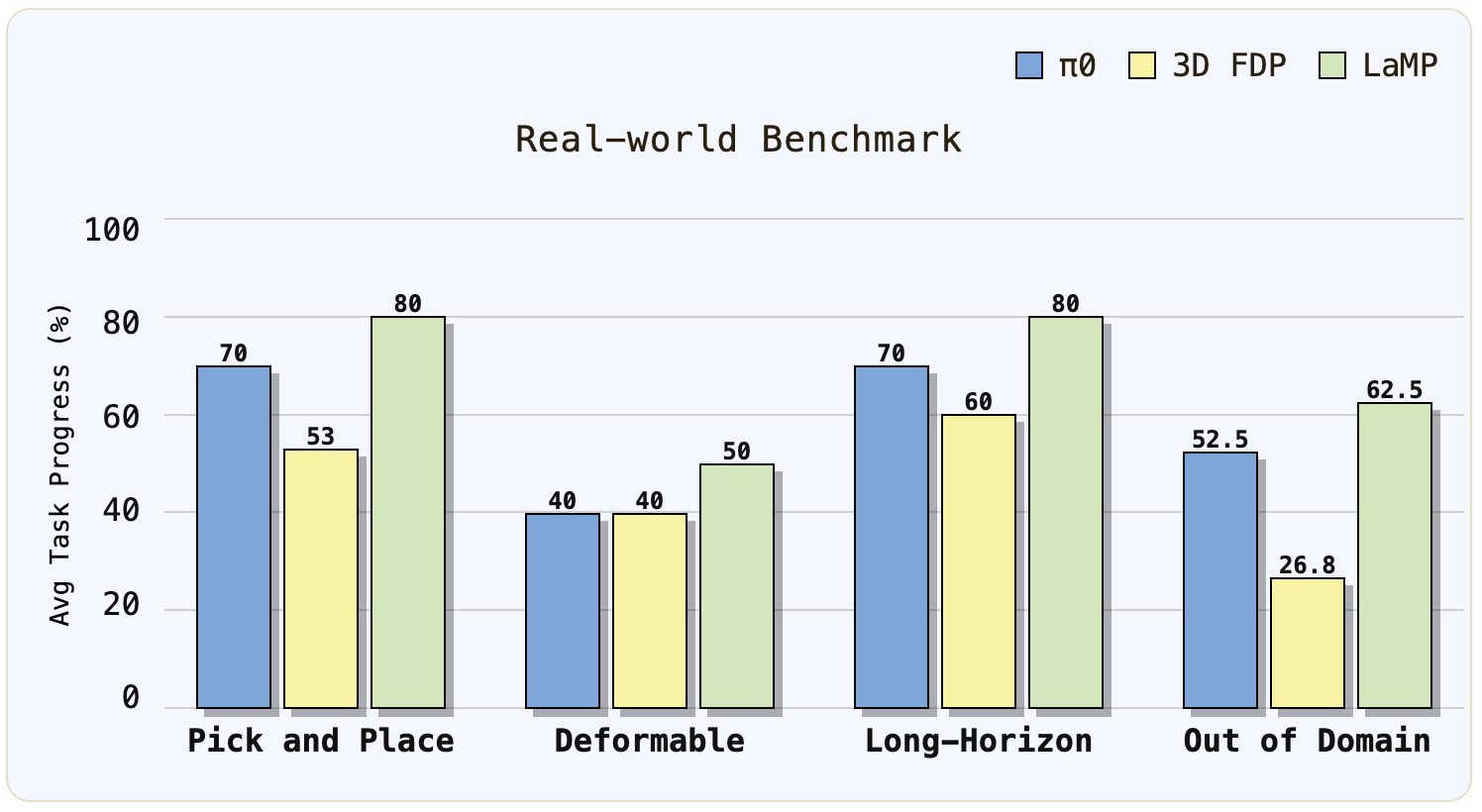}
  \end{center}
  \vspace{-1\baselineskip}
  \caption{\textbf{Real-world task evaluations}. LaMP outperforms $\pi_0$ and 3D FDP across all task categories.}
  \label{fig:real_world_results}
\vspace{-2\baselineskip}
\end{wrapfigure}

\boldparagraph{Baselines.}
We compare LaMP against the following baselines under identical
data budgets:
(1)~$\pi_0$~\cite{pi0}, a mainstream flow-matching-based VLA that directly maps RGB observations to actions;
and (2)~3D Flow Diffusion Policy (3D FDP)~\cite{3dFlowDiffusion}, a diffusion-based
policy that operates on 3D flow representations for
spatially-aware action generation. Since no official implementation is available, we re-implement 3D FDP based on the original paper.

\boldparagraph{Training Protocol.}
We collect 160 expert demonstrations across three tasks on the target robot, totaling approximately one hour of teleoperation data.
Following the warm-up protocol in~\cite{tracegen}, we fine-tune LaMP with 10 demonstrations per task to translate the embodiment-agnostic motion prior into the action space of the target robot, without requiring any robotic action data for pretraining.
We then train a single unified policy on all three tasks using the full dataset, as does $\pi_0$.
Unlike these VLAs, 3D FDP requires training a separate model for each individual task.
This one-policy multi-task design enables cross-task knowledge transfer while preserving the embodiment-agnostic geometric reasoning inherited from the pretrained motion prior.

\boldparagraph{Evaluation Protocol.}
For each in-domain task, we perform 10 trials per method with fixed initial layouts. For
each OOD condition, we perform 5 trials per method per task. Unseen Layout tests randomized object positions and orientations, while Unseen Object and Unseen Background test training-distribution-matched layouts.
We evaluate all methods under identical hardware and lighting conditions to ensure fair comparison. 
We report the average task progress score (see~\cref{fig:platform}) as the primary evaluation metric. 
The metric definition is provided in the supplementary material. 
 
\subsection{Results}
LaMP consistently outperforms both baselines across all task categories (\cref{fig:real_world_results}).
On Pick-and-Place, LaMP achieves \textbf{80\%} average task progress, surpassing $\pi_0$ (70\%) by 10 percentage points and 3D FDP (53\%) by 27 percentage points.
The gains are most pronounced on Deformable manipulation, where LaMP reaches \textbf{50\%} while both $\pi_0$ and 3D FDP plateau at 40\%. This result confirms that the motion prior provides a stronger inductive bias for deformable dynamics.

On Long-Horizon tasks, LaMP maintains \textbf{80\%} progress compared to 70\% for $\pi_0$ and 60\% for 3D FDP, demonstrating superior multi-step planning coherence.
Under Out-of-Domain conditions, LaMP achieves \textbf{62.5\%} while $\pi_0$ drops to 52.5\% and 3D FDP collapses to 26.8\%.
This margin confirms that camera-frame geometric reasoning is more resilient to visual distribution shifts than pixel-level representations.
Notably, 3D FDP fails under OOD conditions, dropping 26.2 points from its in-domain Pick-and-Place performance, whereas LaMP degrades by only 17.5 points.
These results confirm that the embodiment-agnostic motion prior learned from diverse video data transfers effectively to unseen robots without requiring large-scale robotic demonstrations for pretraining.

\subsection{Motion Foresight Visualizations}
We visualize the Motion Expert's predictive foresight to confirm that it captures physically meaningful dynamics rather than serving as opaque regularization.
As shown in \cref{fig:platform} (right), the predicted 3D trajectories anticipate the arm's approach and manipulation \textit{before} execution, spanning rigid grasping, deformable folding, and multi-step long-horizon tasks.
Note that the Action Expert only conditions on 1-step partially denoised hidden states ($\tau=0.1$); the full ODE rollout from $\tau=0$ to $\tau=1.0$ is performed solely for visualization.
\section{Limitations and Future Work}
\label{sec:conclusion}

LaMP demonstrates that dense 3D scene flow serves as an effective latent motion prior within pretrained VLA frameworks, with particular gains in long-horizon planning and cross-embodiment transfer.
Several limitations remain. Our Motion Expert operates at a fixed spatial resolution and temporal horizon, so adaptive-resolution and longer-horizon motion representations are a natural extension. It also depends on TraceForge-generated 3D scene flow for supervision, and learning motion priors directly from unlabeled in-the-wild video would further improve scalability. Finally, motion is injected only at the last VLM layer to avoid the representational collapse observed with multi-layer fusion, and both a full multi-layer ablation and broader real-world, multi-embodiment evaluation are left for future work.
 
\section*{Acknowledgements}
This work was supported by 
the National Natural Science Foundation of China (No. 62506232); 
Shanghai Committee of Science and Technology (Yangfan: No. 24YF2722000; No.24511103200);
Science and Technology Major Project of Jiangsu Province (No. BG2024041).

Lixin Yang (corresponding author) is affiliated with the School of Artificial Intelligence, Shanghai Jiao Tong University and serving as a project advisor at the Shanghai Innovation Institute.

%
%
\bibliographystyle{splncs04}
\bibliography{main}

\newpage
\appendix
\onecolumn
\section{Training and Inference Algorithms}
\label{appendix:algorithms}

We define the notation shared across Algorithm~\ref{alg:lamp_stage2} and Algorithm~\ref{alg:lamp_infer}. $o_t$ is the visual observation at timestep $t$. $l$ is the language instruction. $s_t$ is the robot proprioceptive state. $z$ represents the VLM features extracted from $o_t$ and $l$. $m^0$ is the initial motion noise sampled from a standard Gaussian. $t_1$ is the first timestep of the ODE solver. $z_{\mathrm{m}}$ is the motion hidden state extracted from the Motion Expert after one-step denoising. $z_{\mathrm{guided}}$ is the guided VLM features after applying gated cross-attention between $z$ and $z_{\mathrm{m}}$. The following subsections detail each algorithm with task-specific notation.

\subsection{Stage-2 Motion-Guided Policy Training}

We detail the training procedure for the Action Expert with frozen Motion Expert in Algorithm~\ref{alg:lamp_stage2}.

\textbf{Training-specific symbols.} $a_t$ is the ground-truth action. $\tau_a$ is the action flow-time sampled from $\mathrm{Beta}(1.5, 1.0)$. $\epsilon^a$ is the action noise. $a^{\tau_a}$ is the noisy action at flow-time $\tau_a$. $v_\phi$ is the Action Expert velocity network that predicts the action velocity. $\mathcal{L}_{\mathrm{action}}$ is the flow matching loss for action prediction.

\begin{algorithm}[htbp]
  \caption{Stage-2 Motion-Guided Policy Training}
  \label{alg:lamp_stage2}
  \begin{algorithmic}[1]
    \Require A mini-batch $\mathcal{B}$ from the Vision-Language-Action data containing $(o_t, l, s_t, a_t)$, solver steps $N=10$
    \For{each training iteration}
    \State Sample $(o_t, l, s_t, a_t)\sim\mathcal{B}$
    \State Extract VLM features $z \gets \phi_{\mathrm{vl}}(o_t,l)$ \Comment{Last-layer features}
    \State Sample $m^0 \sim \mathcal{N}(0,I)$ \Comment{Initialize motion noise at $\tau=0$}
    \State $t_1 \gets 1/N$
    \State \textbf{without gradient computation:} \Comment{Motion Expert is frozen in Stage 2}
    \State \quad $z_{\mathrm{m}} \gets f_{\mathrm{mot}}^{t_1}(z, m^0)$ \Comment{Extract motion hidden state from the 1st step}
    \State $z_{\mathrm{guided}} \gets \mathrm{Guide}(z, z_{\mathrm{m}})$ \Comment{Single-layer gated cross-attention}
    \State Sample action flow-time $\tau_a \sim \mathrm{Beta}(1.5,1.0)$ and noise $\epsilon^a \sim \mathcal{N}(0,I)$
    \State $a^{\tau_a} \gets (1-\tau_a)\epsilon^a + \tau_a a_t$
    \State $\hat{u} \gets v_\phi(a^{\tau_a}, \tau_a, z_{\mathrm{guided}}, s_t, t_1)$ \Comment{Action prediction conditioned on state and $t_1$}
    \State $\mathcal{L}_{\mathrm{action}} \gets \|\hat{u}-(a_t-\epsilon^a)\|_2^2$
    \State Update the Guidance module and the Action Expert with $\nabla \mathcal{L}_{\mathrm{action}}$
    \EndFor
  \end{algorithmic}
\end{algorithm}

\subsection{LaMP Inference}

We summarize the inference procedure in Algorithm~\ref{alg:lamp_infer}.

\textbf{Inference-specific symbols.} $a^0$ is the initial action noise. $N$ is the total number of ODE solver steps. $\tau_n$ is the flow-time at step $n$. $v_\phi$ is the Action Expert velocity network. $a^1$ is the final clean action after $N$-step integration.

\begin{algorithm}[htbp]
  \caption{LaMP Inference}
  \label{alg:lamp_infer}
  \begin{algorithmic}[1]
    \Require An observation $o_t$, a language instruction $l$, a robot state $s_t$, solver steps $N=10$
    \State Extract VLM features $z \gets \phi_{\mathrm{vl}}(o_t,l)$ \Comment{Last-layer features}
    \State Sample $m^0 \sim \mathcal{N}(0,I)$
    \State $t_1 \gets 1/N$
    \State Predict partial motion: $z_{\mathrm{m}} \gets f_{\mathrm{mot}}^{t_1}(z, m^0)$ \Comment{1-step denoising to extract hidden}
    \State $z_{\mathrm{guided}} \gets \mathrm{Guide}(z, z_{\mathrm{m}})$
    \State Sample $a^0 \sim \mathcal{N}(0,I)$ \Comment{Initialize action noise at $\tau_a=0$}
    \For{$n=1$ to $N$}
    \State $\tau_n \gets \frac{n-1}{N}$, \quad $\tau_{n+1} \gets \frac{n}{N}$ \Comment{Integrate from $\tau_a=0$ to $\tau_a=1$}
    \State $a^{\tau_{n+1}} \gets a^{\tau_n} + (\tau_{n+1}-\tau_n) \, v_\phi(a^{\tau_n}, \tau_n, z_{\mathrm{guided}}, s_t, t_1)$
    \EndFor
    \State \Return $a^1$ \Comment{Return the clean action data}
  \end{algorithmic}
\end{algorithm}

\section{Training Hyperparameters}

We provide complete training configurations for reproducibility.
\label{appendix:training_details}

\subsection{Motion Expert Pretraining}

We pretrain the Motion Expert on 1.6M observation-language-motion triplets generated via the TraceForge pipeline. The training configuration is detailed in Table~\ref{tab:motion_expert_config}.

\begin{table}[h]
\centering
\caption{\textbf{Hyperparameters for Motion Expert pretraining.}}
\label{tab:motion_expert_config}
\small
\resizebox{0.45\textwidth}{!}{%
\begin{tabular}{l|l}
\toprule
\textbf{Configuration} & \textbf{Value} \\
\midrule
Batch size (per GPU) & 32 \\
Global batch size & 32$\times$16 = 512 \\
Hidden dimension & 1024 \\
Transformer layers & 12 \\
Patch size & $1 \times 2 \times 2$ \\
Grid resolution & $20 \times 20$ \\
Optimizer & AdamW \\
Betas & (0.9, 0.95) \\
Weight decay & $1 \times 10^{-8}$ \\
Learning rate & $2 \times 10^{-4}$ \\
Warm-up steps & 0 \\
Scheduler & cosine$\_$with$\_$min$\_$lr \\
Training epochs & 30 \\
Precision & bfloat16 \\
\bottomrule
\end{tabular}
}
\end{table}

\subsection{Action Expert Training}

Table~\ref{tab:action_expert_config} summarizes the training setup for the Action Expert across different benchmarks.

\begin{table}[t]
\centering
\caption{\textbf{Hyperparameters for Action Expert training.}}
\label{tab:action_expert_config}
\resizebox{\textwidth}{!}{%
\begin{tabular}{l|lll}
\toprule
\textbf{Configuration} & \textbf{LIBERO} & \textbf{SimplerEnv} & \textbf{Real World} \\
\midrule
Batch size (per GPU) & 32 & 32 & 32 \\
Global batch size & 32$\times$16 = 512 & 32$\times$16 = 512 & 32$\times$16 = 512 \\
Action chunk horizon $H$ & 9 & 29 & 15 \\
Image resize & 224$\times$224 & 224$\times$224 & 224$\times$224 \\
Action normalization & on & on & on \\
Data shuffling & on & on & on \\
Optimizer & AdamW & AdamW & AdamW \\
Betas & (0.9, 0.95) & (0.9, 0.95) & (0.9, 0.95) \\
Weight decay & 0.0 & 0.0 & 0.0 \\
Learning rate & $1 \times 10^{-4}$ & $1 \times 10^{-4}$ & $2 \times 10^{-4}$ \\
Warm-up steps & 0 & 0 & 0 \\
Scheduler & cosine$\_$with$\_$min$\_$lr & cosine$\_$with$\_$min$\_$lr & cosine$\_$with$\_$min$\_$lr \\
Training steps & 15k & 15k & 20k \\
Precision & bfloat16 & bfloat16 & bfloat16 \\
\bottomrule
\end{tabular}
}
\end{table}

\section{Task Progress Metric Definition}
\label{appendix:metric}
We define the evaluation protocol for real-world experiments.
For real-world experiments, we report the average task progress score as the primary evaluation metric. We define task progress as the percentage of stages completed within a task episode and average the final progress score across all evaluation trials for each task.

\boldparagraph{Pick-and-Place Tasks.} Each task consists of two stages. Stage I involves approaching and grasping the object. Stage II involves moving the object to the target location and releasing it. We assign 0.5 to the progress score for completing Stage I and 1.0 for completing both stages.

\boldparagraph{Deformable Tasks.} Each task consists of four stages. Stage I involves picking and folding the first corner. Stage II involves picking and folding the second corner. Stage III involves adjusting the fold. Stage IV involves finalizing the deformation into the target configuration. We assign 0.25 per stage completed.

\boldparagraph{Long-Horizon Tasks.} Each task consists of three stages. Stage I involves placing the bread on the plate. Stage II involves grasping the pot, pouring the egg, and returning the pot. Stage III involves placing the egg on top of the bread. We assign approximately 0.33 per stage completed.

\section{Runtime and Efficiency Analysis}
\label{appendix:runtime}
We report real-robot runtime on the Flexiv platform of Sec.~\ref{sec:real_exp} (NVIDIA RTX 4090, batch size 1, averaged over 10 real-robot trials). As shown in Table~\ref{tab:runtime}, the Motion Expert runs a single partial-denoising step (1 of 10) rather than full denoising, adding only $+46.3$\,ms ($1.35\times$) latency and $+1.4$\,GB ($1.15\times$) memory over the motion-free variant. With action chunking ($H{=}16$), the robot receives one 16-step action chunk every $177.1$\,ms, i.e., a ${\approx}5.6$\,Hz chunk rate (${\sim}90$\,Hz effective action rate), meeting the real-time requirements of real-robot deployment.

\begin{table}[h]
\centering
\caption{\textbf{Runtime and memory cost of the Motion Expert.} Measured on the Flexiv platform (RTX 4090, bs\,=\,1, 10-trial average). Relative factors over the motion-free variant are shown in parentheses.}
\label{tab:runtime}
\small
\begin{tabular}{lccc}
\toprule
Variant & Memory (GB) $\downarrow$ & Latency (ms) $\downarrow$ & LIBERO Avg. $\uparrow$ \\
\midrule
w/o motion & 9.6 ($1.00\times$) & 130.8 ($1.00\times$) & 92.4 \\
\textbf{LaMP} & 11.0 ($1.15\times$) & 177.1 ($1.35\times$) & \textbf{98.3} \\
\bottomrule
\end{tabular}
\end{table}

\section{Bimanual Evaluation on RoboTwin 2.0}
\label{appendix:robotwin}
To validate the embodiment-agnostic $(u,v,d)$ representation in a bimanual setting, we evaluate LaMP against $\pi_{0.5}$ and X-VLA on 10 RoboTwin 2.0 tasks~\cite{chen2025robotwin} under both clean and randomized (strong domain randomization) settings. Table~\ref{tab:robotwin} reports the per-task success rates. LaMP attains the best average success under both settings and wins the majority of individual tasks, confirming that the embodiment-agnostic motion prior transfers to bimanual manipulation.

\begin{table}[h]
\centering
\caption{\textbf{Bimanual evaluation on 10 RoboTwin 2.0 tasks.} Per-task success rate (\%) under clean and randomized (Rand., strong domain randomization) settings. \textbf{Bold} denotes the best and \underline{underline} the second best among the three methods for each task and setting.}
\label{tab:robotwin}
\small
\setlength{\tabcolsep}{4pt}
\begin{tabular}{l cc cc cc}
\toprule
\multirow{2}{*}{Simulation Task} & \multicolumn{2}{c}{$\pi_{0.5}$~\cite{pi05}} & \multicolumn{2}{c}{X-VLA~\cite{zheng2025x}} & \multicolumn{2}{c}{LaMP} \\
\cmidrule(lr){2-3} \cmidrule(lr){4-5} \cmidrule(lr){6-7}
& Clean & Rand. & Clean & Rand. & Clean & Rand. \\
\midrule
Blocks Ranking RGB   & \cellcolor{linecolor1}{\underline{92}} & \cellcolor{linecolor1}{\underline{85}} & 83 & 83 & \cellcolor{linecolor2}{\textbf{95}} & \cellcolor{linecolor2}{\textbf{93}} \\
Blocks Ranking Size  & 49 & 26 & \cellcolor{linecolor1}{\underline{67}} & \cellcolor{linecolor2}{\textbf{74}} & \cellcolor{linecolor2}{\textbf{71}} & \cellcolor{linecolor1}{\underline{62}} \\
Handover Mic         & \cellcolor{linecolor2}{\textbf{98}} & \cellcolor{linecolor1}{\underline{97}} & 0 & 0 & \cellcolor{linecolor2}{\textbf{98}} & \cellcolor{linecolor2}{\textbf{98}} \\
Move Can Pot         & 51 & 55 & \cellcolor{linecolor2}{\textbf{89}} & \cellcolor{linecolor2}{\textbf{86}} & \cellcolor{linecolor1}{\underline{68}} & \cellcolor{linecolor1}{\underline{68}} \\
Move Stapler Pad     & 56 & 42 & \cellcolor{linecolor2}{\textbf{78}} & \cellcolor{linecolor2}{\textbf{73}} & \cellcolor{linecolor1}{\underline{70}} & \cellcolor{linecolor1}{\underline{66}} \\
Open Microwave       & 34 & \cellcolor{linecolor1}{\underline{77}} & \cellcolor{linecolor1}{\underline{79}} & 71 & \cellcolor{linecolor2}{\textbf{92}} & \cellcolor{linecolor2}{\textbf{93}} \\
Place Can Basket     & \cellcolor{linecolor1}{\underline{62}} & \cellcolor{linecolor1}{\underline{62}} & 49 & 52 & \cellcolor{linecolor2}{\textbf{81}} & \cellcolor{linecolor2}{\textbf{70}} \\
Place Dual Shoes     & 75 & 75 & \cellcolor{linecolor1}{\underline{79}} & \cellcolor{linecolor1}{\underline{88}} & \cellcolor{linecolor2}{\textbf{84}} & \cellcolor{linecolor2}{\textbf{91}} \\
Place Fan            & \cellcolor{linecolor1}{\underline{87}} & \cellcolor{linecolor1}{\underline{85}} & 80 & 75 & \cellcolor{linecolor2}{\textbf{89}} & \cellcolor{linecolor2}{\textbf{94}} \\
Stack Blocks Three   & \cellcolor{linecolor2}{\textbf{91}} & \cellcolor{linecolor1}{\underline{76}} & 6 & 10 & \cellcolor{linecolor2}{\textbf{91}} & \cellcolor{linecolor2}{\textbf{83}} \\
\midrule
Average              & \cellcolor{linecolor1}{\underline{69.5}} & \cellcolor{linecolor1}{\underline{68.0}} & 61.0 & 61.2 & \cellcolor{linecolor2}{\textbf{83.9}} & \cellcolor{linecolor2}{\textbf{81.8}} \\
\bottomrule
\end{tabular}
\end{table}
\end{document}